\documentclass[runningheads]{llncs}
\usepackage[T1]{fontenc}
\usepackage{cite}
\usepackage{tabularx}
\usepackage{amsmath,amssymb,amsfonts}
\usepackage{algorithmic}
\usepackage{graphicx}
\usepackage{textcomp}
\usepackage{xcolor}
\usepackage{caption}
\usepackage{adjustbox}
\usepackage{booktabs}
\usepackage{wrapfig}
\usepackage[misc]{ifsym}
\newcommand{\method}{FaceSnap}
\begin{document}
\title{FaceSnap: Enhanced ID-fidelity Network for Tuning-free Portrait Customization}
\titlerunning{FaceSnap}
%\author{Anonymous submission}
\author{Benxiang Zhai\inst{1} \and
Yifang Xu\inst{1} \and
Guofeng Zhang\inst{2} \and 
Yang Li\inst{1} \and
Sidan Du\inst{1(}\textsuperscript{\Letter} \inst{)}}

\authorrunning{B. Zhai et al.}
%
%\institute{ }
\institute{Vision AI System Lab, Nanjing University, Nanjing, China\\
\email{zbx@smail.nju.edu.cn, xuyifang123@gmail.com, \{yogo,coff128\}@nju.edu.cn}\and
Research and Development Department, Wonxing Technology, Shanghai, China
\email{zhangguofeng@wonxing.com}}

\maketitle              % typeset the header of the contribution
\begin{abstract}
Benefiting from the significant advancements in text-to-image diffusion models, research in personalized image generation, particularly customized portrait generation, has also made great strides recently. However, existing methods either require time-consuming fine-tuning and lack generalizability or fail to achieve high fidelity in facial details. To address these issues, we propose \method, a novel method based on Stable Diffusion (SD) that requires only a single reference image and produces extremely consistent results in a single inference stage. This method is plug-and-play and can be easily extended to different SD models. Specifically, we design a new Facial Attribute Mixer that can extract comprehensive fused information from both low-level specific features and high-level abstract features, providing better guidance for image generation. We also introduce a Landmark Predictor that maintains reference identity across landmarks with different poses, providing diverse yet detailed spatial control conditions for image generation. Then we use an ID-preserving module to inject these into the UNet. Experimental results demonstrate that our approach performs remarkably in personalized and customized portrait generation, surpassing other state-of-the-art methods in this domain.

\end{abstract}

\section{Introduction}
\label{sec:intro}
Recently, revolutionary breakthroughs have been made in image generation and video understanding \cite{GLIDE-2021, DALL_E2-2022, Imagen-2022, MH-DETR-2024, Moment-GPT-2025, VTG-GPT-2024, GPTSee-2023}, thanks to the development of diffusion-based large text-to-image models such as GLIDE\cite{GLIDE-2021}, DALL-E 2\cite{DALL_E2-2022}, Imagen\cite{Imagen-2022}, SD\cite{SD-2022}, and RAPHAEL\cite{Raphael-2024}. 
The above work has been widely applied in various multimodal fields \cite{PFANet-2021, DAFF-Net-2021, MF-Net-2024, QR-Net-2023}.
A focal point of these developments is personalized and customized portrait generation, which refers to the ability to generate high-fidelity images of an individual, given one or more reference images. This application presents two main challenges: firstly, facial identity preservation requires more comprehensive information, including not only the overall facial structure but also detailed information on various parts of the face; secondly, once the comprehensive information is obtained, it is necessary to determine how to inject it into the UNet\cite{UNet-2015} to better guide image generation.

In order to enhance the effectiveness of personalized and customized portrait generation, many researchers have made significant contributions. These works can be classified into two types based on whether fine-tuning is required during inference. The first type requires fine-tuning during the inference process, including methods such as LORA\cite{LORA-2021}, DreamBooth\cite{Dreambooth-2023}, and Textual Inversion\cite{Texture_Inversion-2022}. This type of approach involves training an adapter or fine-tuning pre-trained text-to-image models based on multiple images during inference. Although these methods achieve high fidelity, they require a significant amount of time for training during inference, making it time-consuming, resource-intensive, and not easy to generalize.  The second type is to generate results directly in one stage without fine-tuning during the inference process \cite{HiFi-Portrait-2025, InstantID-2024, PhotoMaker-2023, HP-3-2025, IP-Adapter-2023, PulID-2024}, including HiFi-Portrait \cite{HiFi-Portrait-2025}, PhotoMaker\cite{PhotoMaker-2023}, IP-Adapter\cite{IP-Adapter-2023},  PuLID\cite{PulID-2024}, InstantID\cite{InstantID-2024} and Diff-PC~\cite{Diff-PC-2025}. This type of approach typically involves using a pre-trained image encoder to extract image features, which are then integrated into the UNet as conditioning image prompts. Although it generally achieves lower fidelity, the single-stage inference approach significantly saves time and computational resources, making it the current mainstream method. Among these methods, while InstantID achieves results comparable to fine-tuning approaches, the extracted image prompt still lacks fine-grained information, leading to less accurate generation of facial details. Additionally, the spatial control provided by the 5-point keypoints is not sufficiently strong, further diminishing the similarity of the generated face.

To address these challenges, we propose \method, a new approach for personalized and customized portrait generation that can produce high-fidelity images in a single inference step, using only a single reference image without the need for fine-tuning. To tackle the issue of obtaining insufficiently detailed and comprehensive facial features, we design a Facial Attribute Mixer that integrates fine-grained CLIP\cite{CLIP-2021} features with abstract face ID embedding\cite{FaceNet-2015} to extract comprehensive fused features. To better control facial shape and pose, we introduce 72-point landmarks as spatial control conditions. Meanwhile, to preserve facial identity under different poses, we design a Landmark Predictor, which can produce landmarks that preserve the identity of the source image while adopting the pose from the driving image, ensuring that the generated images maintain better ID fidelity across various poses. Experimental results demonstrate the superior performance of our proposed method compared to other approaches, achieving state-of-the-art (SOTA) results.

\begin{figure*}[h!]
  \centering
  \includegraphics[width=1.0\linewidth]{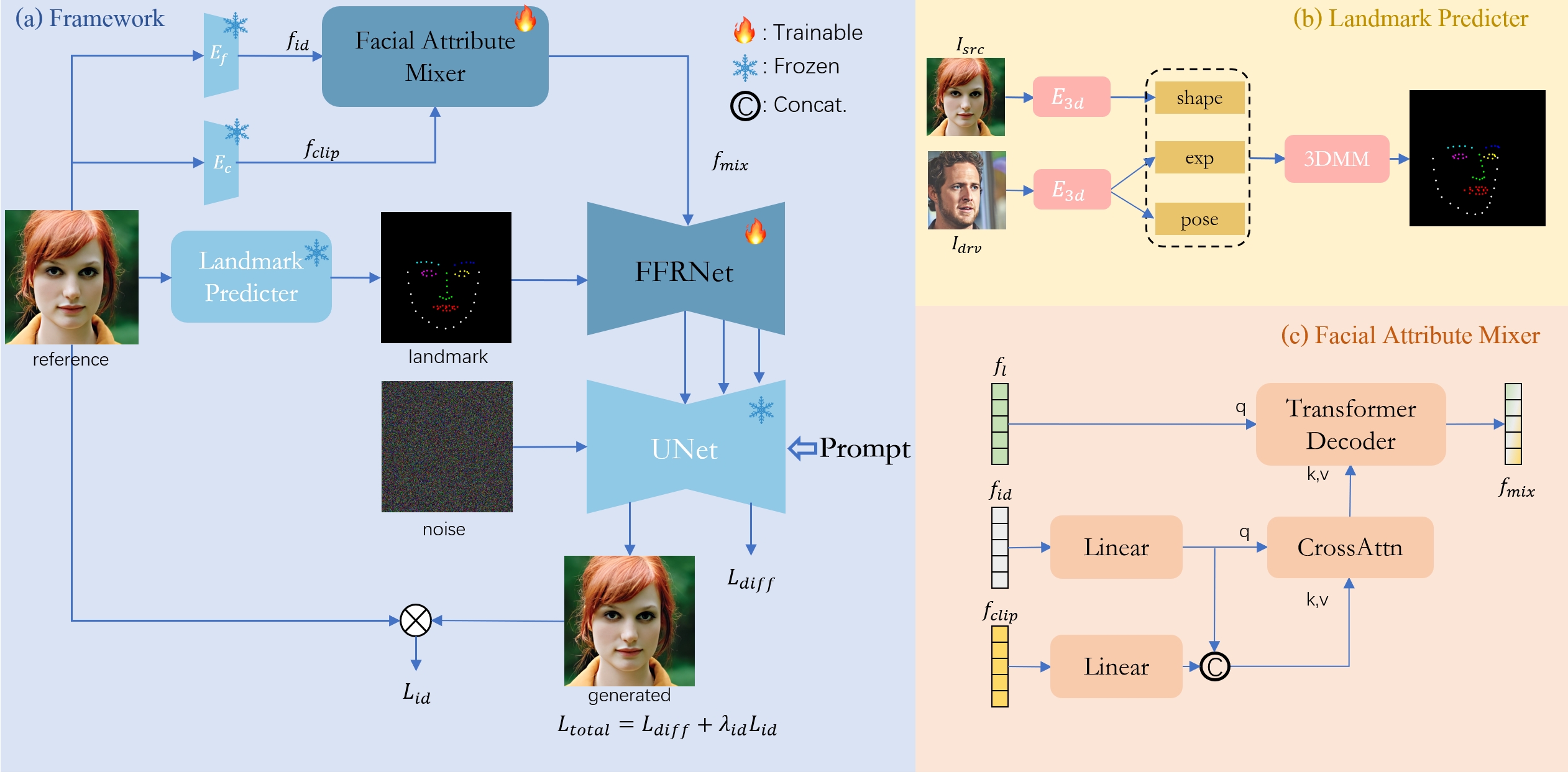}
  \caption{
    The overall framework of our proposed \method. The framework mainly consists of three modules: Facial Attribute Mixer, Face Fidelity Reinforce Network, and Landmark Predictor. We utilize pre-trained models to extract CLIP image features and face ID embeddings. The Facial Attribute Mixer then combines detailed facial information and facial structural information to extract comprehensive fused features. Facial landmarks serve as spatial control conditions, and the Face Fidelity Reinforce Network is introduced to encode both the spatial control conditions and the fused facial features. During inference, to maintain the fidelity of the generated face across different poses, we design the Landmark Predictor to generate landmarks retaining the facial structure of the reference image.
  }
  \label{fig:structure}
\end{figure*}

\section{Method}
\label{sec:method}

\subsection{\method}
\paragraph{Overview}
Fig. \ref{fig:structure}(a) illustrates the main structure of our model. We devise a Facial Attribute Mixer to extract comprehensive and robust facial features. These features are then encoded together with detailed spatial information through a Face Fidelity Reinforce Network (FFRNet) and integrated into the UNet.

\paragraph{Facial Attribute Mixer}
Previous works either use CLIP\cite{CLIP-2021} image features or face ID embeddings. Since CLIP is trained on a diverse set of images with weak alignment, the features encoded by CLIP are relatively low-level and specific. In contrast, the face model used for face recognition is trained on facial datasets through a clustering approach, resulting in features that are higher-level and abstract. The face model encodes features that more accurately reflect the identity of the individual. However, it lacks certain facial details, which is precisely what CLIP features can complement. Therefore, we design a Facial Attribute Mixer to effectively integrate these two sets of features.

Fig. \ref{fig:structure}(c) shows the structure of the Facial Attribute Mixer. Firstly, we utilize a pre-trained face detection model to crop the facial region and extract face ID embeddings and CLIP features through a pre-trained face model\cite{ArcFace-2019} and CLIP image encoder respectively. Then we use linear layers $\pi_{1}$ and $\pi_{2}$ to map $f_{id}$ and $f_{clip}$ to the same dimension $d$, where $f_{id}^{'} = \pi_{1}(F_{id}) \in \mathbb{R}^{N \times 20 \times d}$ and $f_{clip}^{'} = \pi_{1}(F_{clip}) \in \mathbb{R}^{N \times 257 \times d}$ and $f_{id}$ is mapped to $20$ tokens. After that, we employ $f_{id}^{'}$ as $query$, and concatenate $f_{id}^{'}$ and $f_{clip}^{'}$ as $key$ and $value$ to perform cross-attention calculations to obtain the preliminary fused features $f_{pre} \in \mathbb{R}^{N \times 20 \times d}$. Finally, a transformer decoder is used to extract the final fusion features $f_{mix}\in \mathbb{R}^{N \times 16 \times d}$, where the preliminary fusion features $f_{pre}$ are utilized as the $key$ and $value$, while learnable queries $f_{l}\in \mathbb{R}^{N \times 16 \times d}$ are input as $query$.  The above process can be briefly represented as follows:
\begin{gather}
    f_{pre} = \text{CrossAttn}(q=f_{id}', kv=concat(f_{id}',f_{clip}')) \\
    f_{mix} = \text{Decoder}(q=f_{l},kv=f_{pre})
\end{gather}

\paragraph{Face Fidelity Reinforce Network}
We use a ControlNet to inject the comprehensive fused features $f_{mix}$ into the UNet. For the control image input into ControlNet, we use 72-point landmarks, which not only enable spatial control but also contribute to ID preservation. Following previous works \cite{DECA-2021, Face2FaceRHO-2022}, we design a Landmark Predictor to better maintain identity when generating images with different poses. As shown in the Fig. \ref{fig:structure}(b), given a source image whose identity we aim to preserve and a driving image providing pose information, we utilize the 3D reconstruction network, DECA \cite{DECA-2021} to extract the facial shape, pose, and expression information \begin{math}[{s_{s},p_{s},e_{s}}],[{s_{d},p_{d},e_{d}}]\end{math} from the two images, respectively. We then combine the facial shape information from the source image with the pose and expression information from the drive image \begin{math}[{s_{s},p_{d},e_{d}}]\end{math} to render a new 3D face using the 3D Morphable Model (3DMM) fitting algorithm. Finally, we project the 3D face onto the 2D plane to predict the landmarks. In this manner, we achieve the desired pose while effectively preserving the identity of the source image.

Additionally, instead of using text prompts, we use the comprehensive fused features $f_{mix}$ as conditions for the cross-attention layers. On one hand, recent works\cite{T2I-Adapter-2024, SCEdit-2024, ControlNext-2024} have shown that the spatial control effectiveness of ControlNet does not necessarily require text prompts. On the other hand, we aim to encourage ControlNet to focus more on identity information and effectively inject it into the UNet.

\subsection{Training and Inference}
During training, we optimize the Facial Attribute Mixer and FFRNet while keeping the parameters of the pre-trained SD model frozen. We train the model on facial datasets with image-text pairs, and employ masked diffusion loss\cite{masked-diffusion-loss-2023} to ensure the model focuses on the facial region without being influenced by background biases:
\begin{equation}
\label{eq:masked_diff_loss}
\begin{aligned}
    &L_{diff} = 
    &E_{z,y,c,\epsilon\sim\mathcal{N}(0,1),t}[\parallel\epsilon\odot M_{f}-\epsilon_{\theta}(z_{t},t,\tau_{\theta}(y),c)\odot M_{f}\parallel_{2}^{2}]
\end{aligned}
\end{equation}
where \begin{math}{z_{t}}\end{math} is the noise latent, \textit{y} is the text prompt, \begin{math}\tau_{\theta}\end{math} is the frozen CLIP text encoder, \textit{t} is the timestep, \begin{math}\epsilon\end{math} is sampled from the standard Gaussian distribution, and \begin{math}\epsilon_{\theta}\end{math} is the Unet, \textit{c} is the image condition for the \method, \begin{math}{M_{f}}\end{math} is the facial region mask.

Moreover, inspired by PuLID\cite{PulID-2024}, to enhance the focus on ID preservation during training, we also use a lightning branch\cite{SDXL_Lightning-2024} to quickly generate accurate images and calculate ID loss by comparing them with reference images:
\begin{equation}
\label{eq:ID_loss}
L_{id}=1-CosSim(\phi(C_{id}),\phi(\mathrm{L\text{-}T2I}(x_{T},C_{id},C_{txt})))
\end{equation}
where \begin{math}C_{id}\end{math} is the reference image, \begin{math}C_{txt}\end{math} means text prompt,  \begin{math}x_{T}\end{math} denotes the pure noise, L-T2I represents the Lightning T2I branch, and \begin{math}\phi\end{math} refers to the face recognition backbone.
The full learning objective is defined as:
% reconstruction loss + ID loss
\begin{equation}
\label{eq:total_loss}
    L_{total} = L_{diff} + \lambda_{id}L_{id}
\end{equation}
where \begin{math}\lambda_{id}\end{math} serves as a hyperparameter that determines the relative importance of each loss item.

\begin{figure*}[t]
  \centering
  \includegraphics[width=1.0\linewidth]{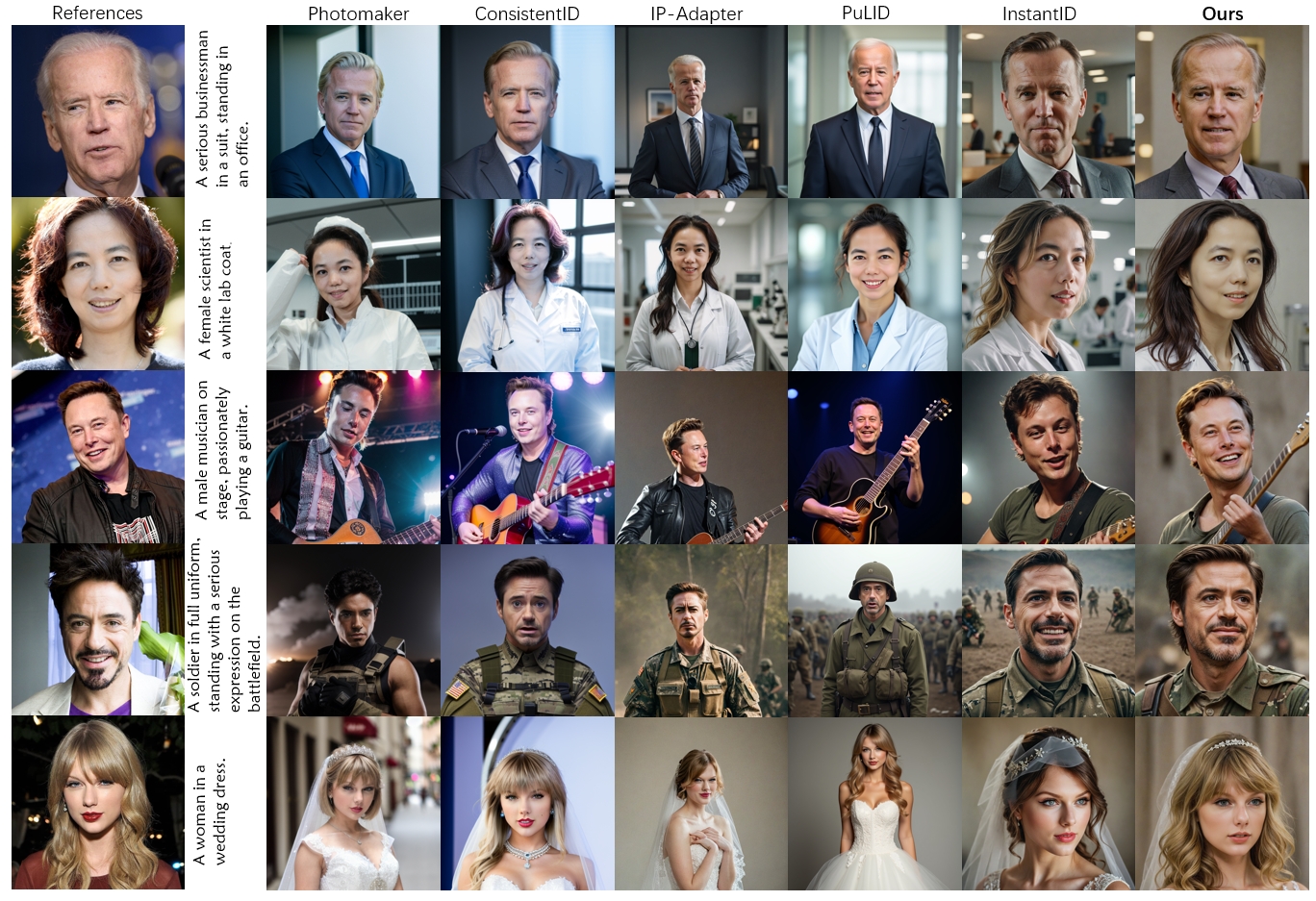}
  \caption{Qualitative comparison samples. We compare our \method\ with ConsistentID\cite{ConsistentID-2024}, Photomaker\cite{PhotoMaker-2023}, IP-Adapter\cite{IP-Adapter-2023}, PuLID\cite{PulID-2024} and InstantID\cite{InstantID-2024} using five distinct identities with corresponding prompts. It can be observed that our method has the ability to generate higher-quality and fidelity images.}
\label{fig:visual}
\end{figure*}

\begin{table}
\caption{Quantitative comparison on the universal recontextualization setting. The benchmark metrics evaluate text consistency (CLIP-T), the preservation of ID information (CLIP-face, FaceSim) and the generation quality (FID). Additionally, the inference time (Time) and memory consumption (VRAM) are also presented.}
\begin{tabularx}{\textwidth}{p{2.6cm}>{\centering\arraybackslash}p{1.5cm}>{\centering\arraybackslash}p{2cm}>{\centering\arraybackslash}p{1.5cm}>{\centering\arraybackslash}X>{\centering\arraybackslash}X>{\centering\arraybackslash}X}
\hline
             & \textbf{CLIP-T$\uparrow$} & \textbf{CLIP-face$\uparrow$} & \textbf{FaceSim$\uparrow$} & \textbf{FID$\downarrow$} & \textbf{Time$\downarrow$} 
             & \textbf{VRAM$\downarrow$}\\ 
\hline
Photomaker\cite{PhotoMaker-2023}      & 30.5& 74.6& 61.7& 226.3& 4.6s& \textbf{15.1GB} \\
ConsistentID\cite{ConsistentID-2024}  & 31.1& 77.4& 64.8& 219.4& 5.2s& 17.7GB \\
IP-Adapter\cite{IP-Adapter-2023}      & \textbf{33.2}& 76.7& 67.6& 207.5& \textbf{4.4s}& 17.2GB  \\
PuLID\cite{PulID-2024}                & 33.0& 78.7& 72.5& 208.6& 4.9s& 17.4GB\\
InstantID\cite{InstantID-2024}        & 32.8& 78.9& 71.1& 208.9& 5.4s& 19.8GB \\
\method                           & 32.6& \textbf{81.4}& \textbf{73.6}& \textbf{205.6}& 6.1s& 18.1GB  \\ 
\hline
\end{tabularx}
\label{tab:quantitative_comparison}
\end{table}

\section{Experiments}
\label{sec:experiments}

\subsection{Experimental Settings}
\label{subsec:settings}

\paragraph{Implementation details}
For the training data, we collected 160,000 facial images from VGGFace2\cite{dataset-VGGFace2-2018}, FFHQ\cite{dataset-FFHQ-2019}, and CelebA-HQ\cite{dataset-celebahq-2017} datasets. To enhance the model's ability to generate portraits at various scales, we apply augmentations such as scaling, rotation, and padding. Ultimately, we created a dataset containing 800,000 images, with corresponding captions generated by BLIP2\cite{BLIP2-2023}. To achieve better generative performance, we used SDXL\cite{SDXL-2023} as the base model. We first cropped the facial regions from the images and then used CLIP-ViT-H\cite{CLIP-2021} as the image encoder to extract image features, while a pre-trained face model\footnote{https://github.com/deepinsight/insightface} was employed to extract face ID embeddings. \method\ is trained on 2 NVIDIA A800 GPUs (80G) with a batch size of 7 per GPU for 360000 steps. We use the AdamW\cite{AdamW-2017} optimizer and set the learning rate to 1e-5. During training, to improve the generation performance by using classifier-free guidance\cite{CFG-2022}, we set a 10\% chance of replacing the original text embedding with a null text embedding. 
In the inference stage, to save time, we used DreamShaperXLv2.1 Turbo\footnote{https://civitai.com/models/112902/dreamshaper-xl} as the image generation model, employed the DPM++ SDE\cite{KPS++SDE-2022} sampler, and set the guidance scale to 2 and \begin{math}\lambda_{id}\end{math} to 0.5.

\paragraph{Evaluation metrics} 
To evaluate the generation quality of \method, we used four widely adopted metrics: CLIP-face\cite{Texture_Inversion-2022}, FaceSim\cite{FaceNet-2015}, CLIP-T\cite{CLIP-2021}, and FID\cite{FID-2017}. CLIP-face measures the average cosine similarity between the CLIP embeddings of real and generated images. FaceSim measures the similarity between face ID embeddings extracted using FaceNet. These two metrics evaluate ID fidelity from two different perspectives. CLIP-T calculates the average pairwise cosine similarity between text embeddings and image CLIP embeddings, assessing the degree of text alignment. Meanwhile, FID is used to evaluate the quality of the generation. Time and VRAM refer to the inference time and memory consumption, respectively, for generating a 1024×1024 image with 8 sampling steps on a single A800 GPU using FP16 precision.

\subsection{Comparisons}
\label{subsec:comparison}
To demonstrate the effectiveness of \method, we conduct a comprehensive comparison with existing state-of-the-art methods. For these methods, we use the officially provided models and default parameters.
\paragraph{Qualitative results}
To visually demonstrate the superiority of \method, we present the generation results of different methods using five different identities under various poses and prompts. As shown in Fig. \ref{fig:visual}, the visualization results indicate that \method\ can generate high-fidelity, realistic images while aligning well with the provided text prompts, surpassing other methods.

\paragraph{Quantitative results}
We collected 200 ID images from the Internet which were not used in training and serve as the validation set. To comprehensively evaluate the ID-preserving generation capability of these methods, we provide 10 different pose templates varying in head orientation and head size. The results are shown in Tab. \ref{tab:quantitative_comparison}. \method\ significantly outperforms other methods in both the FaceSim and CLIP-face metrics, demonstrating its superiority in ID preservation. This can be attributed to the Facial Attribute Mixer’s ability to extract identity information and the landmark-based spatial control that enhances identity retention. Although \method\ does not achieve the best result in the CLIP-T metric, this could be due to the model’s focus on ID preservation over alignment with the text prompt. However, the slightly lower performance compared to the SOTA methods still demonstrates the effectiveness of our method.

\paragraph{Deployment feasibility}
Similar to other approaches, \method\ is capable of running on any GPU with more than 24GB of VRAM.

\begin{figure}[ht]
\centering
\includegraphics[width=\linewidth]{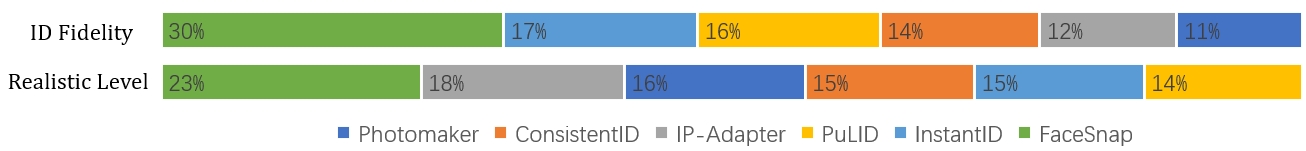}
\caption{User preferences on realistic level, ID fidelity for different methods}
\label{fig:user_study}
\end{figure}

\paragraph{User study.}
To provide a more objective evaluation of \method, we conduct a user survey where participants are asked to select the method that produced the best images in terms of realism and ID preservation. We randomly select 10 IDs(5 prompts, 5 poses) and generate results using each method. Forty users rate the images based on ID fidelity and realistic level. As shown in Fig. \ref{fig:user_study}, the results indicate that our method received the highest user preference.

\begin{figure*}[h]
  \centering
  \includegraphics[width=1.0\linewidth]{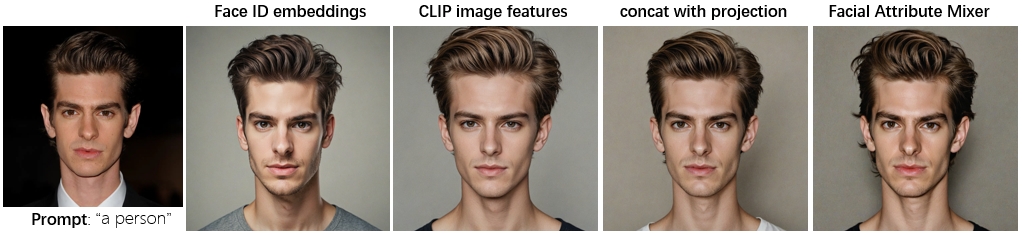}
  \caption{\small Impact of Facial Attribute Mixer. Compared with "concat with projection", using the Facial Attribute Mixer to obtain facial features effectively captures the ID details from Face ID embeddings and CLIP image features, thereby enhancing ID fidelity.}
\label{fig:face_mixer}
\end{figure*}

\begin{table}[h]
\caption{Ablation study on the effect of Facial Attribute Mixer.}
\begin{tabularx}{\textwidth}{l>{\centering\arraybackslash}X>{\centering\arraybackslash}X>{\centering\arraybackslash}X}
\hline
    & \textbf{CLIP-face$\uparrow$} & \textbf{FaceSim$\uparrow$} & FID$\downarrow$ \\ 
\hline
Face ID embeddings & 74.4 & 67.2 & 216.9\\
CLIP image features & 76.4 & 66.1 & 217.2 \\
concat with projection & 77.6 & 68.8 & 214.8 \\
\textbf{Facial Attribute Mixer} & \textbf{79.2} & \textbf{70.6} & \textbf{212.3}\\ 
\hline
\end{tabularx}
\label{tab:face_mixer}
\end{table}

\subsection{Ablation Studies} 
\paragraph{Impact of Facial Attribute Mixer}
We also conducted an ablation study to evaluate the impact of the Facial Attribute Mixer. We established the following three scenarios: using only the CLIP image features, the face ID embeddings, and both features with only a projection layer. As shown in Tab.\ref{tab:face_mixer}, using only CLIP image features improves the ID consistency of facial details, as indicated by the increase in the CLIP-face value. When using only face ID embeddings, the results are the opposite; it enhances face similarity but lacks the ability to preserve fine facial details. Combining both features balances the model's performance. However, using only a single projection layer leads to insufficient feature extraction and fusion. 
In contrast, our proposed Facial Attribute Mixer achieves the best results in both metrics, demonstrating that it effectively fuses the two features. Fig.\ref{fig:face_mixer} shows the visual results under various facial feature configurations.

\begin{figure}[h]
  \centering
  \includegraphics[width=1.0\linewidth]{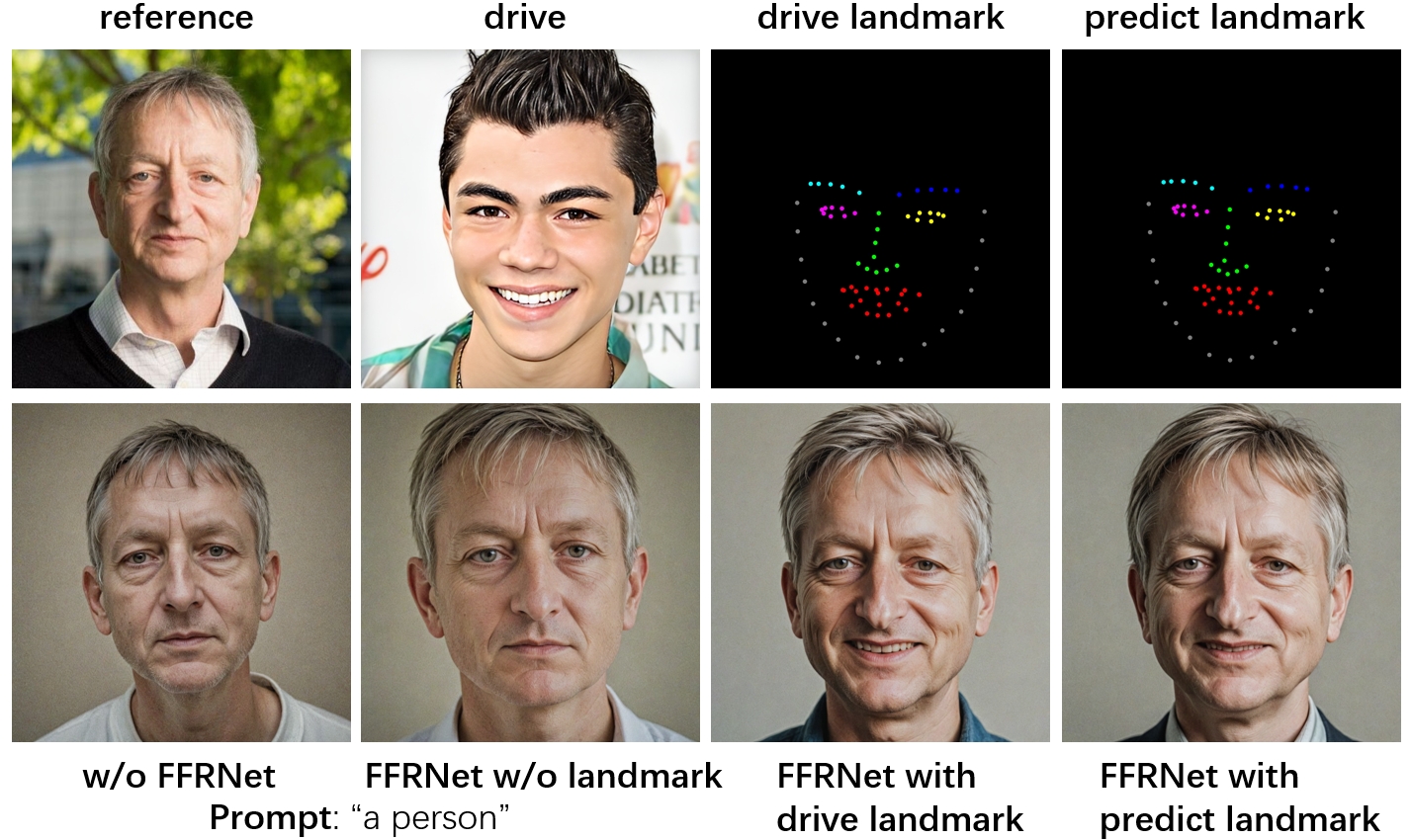}
  \caption{\small Effect of FFRNet and Landmark Predictor. FFRNet and landmarks used as spatial control conditions help with ID fidelity. When the face shapes of the reference and driving images differ significantly, the Landmark Predictor generates landmarks that reflect the face shape of the reference image, thereby improving ID fidelity.}
\label{fig:FFRNet}
\end{figure}

\begin{table}[h]
\caption{Ablation study on the impact of FFRNet and Spatial Control Conditions.}
\begin{tabularx}{\textwidth}{l>{\centering\arraybackslash}X>{\centering\arraybackslash}X>{\centering\arraybackslash}X}
\hline
    & \textbf{CLIP-face$\uparrow$} & \textbf{FaceSim$\uparrow$} & FID$\downarrow$\\
\hline
without FFRNet & 70.2 & 59.8 & 225.2\\
FFRNet without landmark & 72.6 & 61.3 & 218.5\\
FFRNet with drive landmark & 77.8 & 69.3 & 213.7\\
FFRNet with Landmark Predicter & \textbf{79.2} & \textbf{70.6} & \textbf{212.3}\\
\hline
\end{tabularx}
\label{tab:ablation}
\end{table}

\label{subsec:ablation}
\paragraph{Effect of FFRNet and Spatial Control Condition}
We examined the impact of FFRNet and spatial control conditions on the generation results. To validate the effectiveness of FFRNet, we concatenated the comprehensive fusion features extracted by the Facial Attribute Mixer with text features, and embedded them into the pre-trained cross-attention layers. To validate the effect of spatial control conditions, we designed two schemes: one without landmarks, where a black image was input to the FFRNet, and another where we did not use our designed Landmark Predictor but directly extracted landmarks from the drive image which provides the pose. From the results in Tab. \ref{tab:ablation}, we observed that using FFRNet significantly improves facial similarity, and employing landmarks as spatial control conditions also contributed to an increase in facial similarity to some extent. Comparing the results with directly extracting landmarks from the drive image highlights the effectiveness of our designed Landmark Predictor in terms of ID fidelity. Fig.\ref{fig:FFRNet} shows the visual results of the effect of FFRNet and Landmark Predictor.

\section{Conclusion}
In this paper, we introduce \method, a novel zero-shot personalized and customized portrait generation method that requires only a single reference image and can control the generated results using an additional drive image. We design the Facial Attribute Mixer, which effectively extracts comprehensive fused features from CLIP image features and face ID embeddings. We introduce the FFRNet to integrate facial features and spatial control conditions and design a Landmark Predictor to generate landmarks that preserve facial shape, providing robust spatial control conditions. Quantitative and qualitative experimental results demonstrate the superior stability and accuracy of our method, significantly surpassing other approaches in terms of ID fidelity.

Despite the effectiveness of our \method, there remain areas for improvement. The high parameter count of ControlNet results in substantial memory consumption and increased inference time. Recent studies have shown that existing ControlNet architectures may include redundant components. Therefore, exploring ways to streamline ControlNet’s structure could lead to enhanced efficiency. Additionally, although our methodachieves high-quality generation using just a single reference image, multiple reference images inherently contain more comprehensive information. Simple averaging may not effectively represent all the image features. Therefore, future work should focus on developing a method to fuse features from multiple images of the same identity to obtain a more holistic representation.
\bibliographystyle{splncs04}
\bibliography{icann25}
\end{document}